# Small Drone Field Experiment:
# Data Collection & Processing

**Dalton Rosario, Christoph Borel, Damon Conover**
U.S. Army Research Laboratory
THE UNITED STATES OF AMERICA

dalton.s.rosario.civ@mail.mil, christoph.borel-donehue.civ@mail.mil, damon.m.conover.civ@mail.mil

**Ryan McAlinden**
University of Southern California - Institute for Creative Technologies
THE UNITED STATES OF AMERICA

mcalinden@ict.usc.edu

**Anthony Ortiz**
University of Texas - El Paso
THE UNITED STATES OF AMERICA

amortizcepeda@miners.utep.edu

**Sarah Shiver**
University of California - Santa Barbara
THE UNITED STATES OF AMERICA

swshivers@umail.ucsb.edu

**Blair Simon**
Headwall Photonics
THE UNITED STATES OF AMERICA

bsimon@headwallphotonics.com

*ABSTRACT*

*Following an initiative formalized in April 2016—formally known as ARL West—between the U.S. Army Research Laboratory (ARL) and University of Southern California's Institute for Creative Technologies (USC ICT), a field experiment was coordinated and executed in the summer of 2016 by ARL, USC ICT, and Headwall Photonics. The purpose was to image part of the USC main campus in Los Angeles, USA, using two portable COTS (commercial off the shelf) aerial drone solutions for data acquisition, for photogrammetry (3D reconstruction from images), and fusion of hyperspectral data with the recovered set of 3D point clouds representing the target area. The research aims for determining the viability of having a machine capable of segmenting the target area into key material classes (e.g., manmade structures, live vegetation, water) for use in multiple purposes, to include providing the user with a more accurate scene understanding and enabling the unsupervised automatic sampling of meaningful material classes from the target area for adaptive semi-supervised machine learning. In the latter, a target-set library may be used for automatic machine training with*

*data of local material classes, as an example, to increase the prediction chances of machines recognizing targets. The field experiment and associated data post processing approach to correct for reflectance, geo-rectify, recover the area's dense point clouds from images, register spectral with elevation properties of scene surfaces from the independently collected datasets, and generate the desired scene segmented maps are discussed. Lessons learned from the experience are also highlighted throughout the paper.*

## 1.0 INTRODUCTION

Today's commercially off-the-Shelf (COTS) Unmanned Aerial Vehicles (UAVs) are exceptionally small and light and they demand payloads to match. Before the rapid availability and adoption of today's UAVs, hyperspectral (HS) remote sensing was largely the domain of satellites and maned aircraft. But these platforms can be financially inefficient for the kind of repeatable and continual land surveys, or surveillance and reconnaissance, needed to spot trends and develop solutions to problems. HS sensing instruments are evolving along a parallel track with UAV; both a getting smaller, lighter, and easier to use. The basic function of a HS sensor is to capture individual slices of an incoming scene (through a physical slit) and to break each slice into discrete wavelength components onto a focal plane array (FPA). A diffraction grating manages the task of dispersing the image slices into discrete wavelength components. Ancillary instrumentation such as LiDAR (light detection and ranging) and GPS (global positioning systems) are often joined with HS sensors as elements of the payload suite to help assure that the collected imagery is both complete and precise. Ortho-rectification is a by-product of the outputs from all three. These portable COTS aerial drone solutions for data acquisition are further augmented by the fact that remote sensing from small UAS can map terrain features of potential conflict areas (e.g., dense urban areas) and classify major material classes (e.g., manmade structure, live vegetation, water) to provide enhanced situation awareness.

On that note, following an effort formalized in April 2016, which became known as *ARL West*, between the U.S. Army Research Laboratory (ARL) and University of Southern California's Institute for Creative Technologies (USC ICT), a field experiment was coordinated and executed in the summer of 2016 by ARL, USC ICT, and Headwall Photonics to image part of the USC main campus in Los Angeles, using portable COTS (commercial off the shelf) aerial drone solutions for data acquisition, for photogrammetry—3D reconstruction from images, and fusion of hyperspectral data with the recovered set of 3D point clouds, representing the target area, for applications of interest to the NATO military sensing community. ARL West aims for co-locating Army research and development personnel and gain access to subject matter experts and technical centers and universities not well represented on the U.S. east coast [1].

In the field experiment at USC main campus, the research teams, see Fig. 1, deployed two drones—a quadcopter operated by ICT and a hexacopter operated by Headwall—in addition to a number of calibration target materials and optical sensors; to include color cameras, a handheld spectrometer, and two hyperspectral imagers. Both aerial and ground based data acquisitions were conducted. The airborne data were geo-rectified, a 3D model of the campus was photogrammetrically reconstructed using images from the ICT drone's color camera, and the Headwall drone's hyperspectral data were radiometrically calibrated using ground calibration targets. Ground hyperspectral data were also radiometrically calibrated and qualitatively compared against aerial hyperspectral data of the USC spatial areas that applied. ARL researchers, who supervised and worked with graduate summer interns at ICT, aimed for demonstrating the value of fusing 3D point clouds and hyperspectral datasets produced by COTS emerging small drone/sensing technologies for improved 3D scene segmentation, target material detection, and machine learning. Task-specific, geo-rectified maps—representing segmented regions of the scene's live vegetation materials (e.g., trees, grass), high manmade structures (e.g., buildings), water, and target material classification—produced in this work could feed into the ARL-developed Sensor 3D Common Operating Picture (COP) that was recently employed in a joint field exercise with Defence Research Development Canada (DRDC): Enterprise Challenge 2016 (EC-16). EC-16 was also conducted in July and August 2016, where ARL participated with DRDC to conduct further experimentation and demonstrations incorporating an Expeditionary Processing, Exploitation and Dissemination (Ex-PED) model, as a utility of tactical wide-area and persistent sensing in a bandwidth constrained environment. Canadian assets were fully integrated with the inclusion of the 3D COP to enable appropriate sensor management. ARL's participation in EC-16 is discussed in a separate NATO SET 241 paper [2].

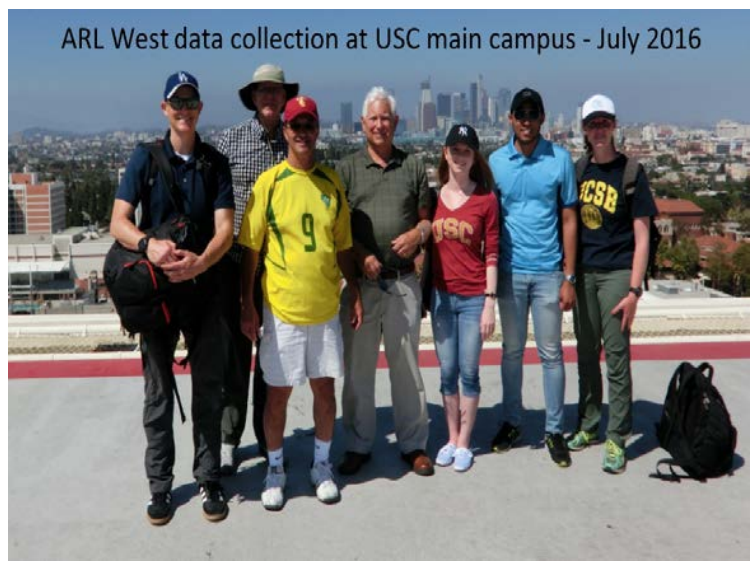

**Figure 1: Aerial and ground data collection at USC main campus. Left to right: Ryan McAlinden (USC/ICT), Christoph Borel (ARL), Dalton Rosario (ARL), Kirk Grim (Headwall Photonics), Britany Beidleman (USC), Anthony Ortiz (U. Texas El Paso), Sarah Shiver (UC Santa Barbara). Not shown in picture are Blair Simon (Headwall) and Bryan Baker (Leica Geosystems).**

This paper also describes a work-flow to leverage these assets for rapid development of simulation-ready terrain assets from data captured in the field. This portion of the effort builds on earlier work in this area [3] and describes ICT current pipeline and its applications in simulation and training both within the research and development communities, as well as operational use cases. We describe the capability for small units (fire teams, squads) with an organic asset such as handheld and launched UAS that allows them to define an Area of Interest (AOI), autonomously fly the aircraft to capture the required imagery, reconstruct the terrain features including elevation, structures, vegetation, and leverage this data for mission rehearsal, simulation, and enhancing situational awareness.

This paper is organized as follows: Section 2 describes the ground and aerial field experiment set up for the hyperspectral data collection portion of the experiment, to include the UAV and remote sensing specifications. Section 3 describes the data collection for photogrammetry. Section 4 describes the fusion of 3D DEM and HS data. Section 5 concludes the paper.

## 2.0 AERIAL & GROUND HYPERSPECTRAL DATA COLLECTION

For the USC main campus aerial hyperspectral data collection, we employed an integrated multi-rotor small UAV and VNIR hyperspectral imaging system with full software control. Details and specifications of the hyperspectral imager in this experiment (Headwall Photonics' Nano-Hyperspec) and the small UAV (Leica Geosystems' Aibotix X6) are shown in Subsection 2.1.1. A ground based hyperspectral imager (Surface Optics Corporation's SOC 710) was also employed in order to cross reference the radiometrically calibrated spectral data of specific material types between two independent hyperspectral imagers operating in the same region of the spectrum; details and specs are shown in Subsection 2.1.2.

### 2.1 Hyperspectral Imaging Technologies

#### 2.1.1 Small Drone and Hyperspectral Imaging System

The employed small drone and remote sensing integrated package combines Headwall's Nano-Hyperspec® VNIR sensor [4] with the Leica-Geosystem Aibotix X6 multi-rotor UAS and advanced GPS technology, as shown in Fig. 2.

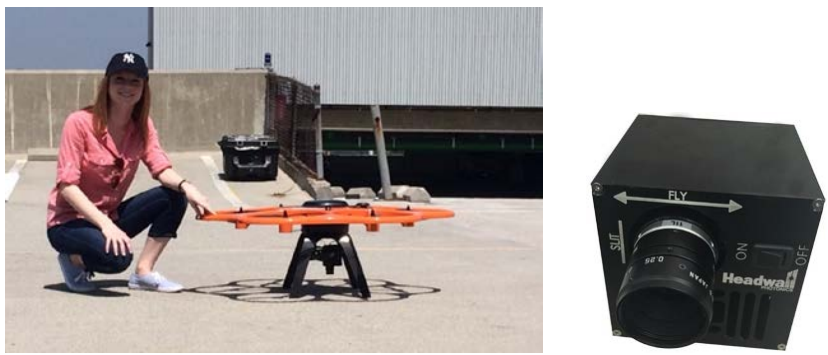

**Figure 2: Headwall's Nano-Hyperspec® VNIR imager (right) integrated package with Leica-Geosystem's Aibotix X6 small drone (left) at USC main campus.**

Headwall's Nano-Hyperspec sensor covers the VNIR range of 400-1000nm with aberration-corrected imaging performance, very high spatial and spectral resolution, and a wide field of view. The stability of the Leica X6 UAV is known for delivering safety and performance across a wide range of deployment scenarios. The Nano-Hyperspec sensor attaches easily to the stabilized gimbal on the X6, assuring stable imaging performance while aloft. A key advantage of Nano-Hyperspec is that it also includes 480GB of on-board data collection/storage, plus attached GPS/IMU functionality. By precisely managing parameters such as flight altitude, speed, direction and position of the sensor, the Leica X6 is suited for the collection of hyperspectral data. The Nano-Hyperspec contains on-board high-speed data-collection capabilities, which makes the integrated package lighter and more stable for increased flight duration. In addition, Headwall's airborne Hyperspec® III software manages key tasks such as post-processing and orthorectification.

Key specifications of the Nano-Hyperspec sensor used in this data collection: 640 spatial bands, 270 spectral bands, rrame rate 300 Hz max, 2.2nm dispersion per pixel (nm/pixel), 6nm FWHM Slit Image, 17mm lens, 480GB storage capacity (~ 130 minutes at 100 fps), connectivity using Gigabit Ethernet, size (exclusive of GPS): 3" x 3" x 4.72" (76.2mm x 76.2mm x 119.92mm), weight without lens: less than 1.15 lb (0.52kg).

The Nano-Hyperspect also uses a concentric imager design that features aberration correction technology that yields an outstanding spatial and spectral resolution, a wide field of view, and high SNR. A wide field of view with aberration correction means that swath widths are maximized and flight time is minimized. By integrating everything in one package, this airborne solution helps save battery life allowing for more time aloft gathering data. In addition, the integrated data-collection system on Nano-Hyperspec has a Gig-E connection that permits quick and easy off-loading of HSI data between flights while synchronized GPS/INS data collection allows for orthorectification during post-processing. The Nano-Hyperspec also uses its own diffraction gratings for outstanding optical performance. Benefits include high signal-to-noise and low stray light in a small-form-factor instrument with no moving parts. This means a high degree of robustness and environmental stability.

### 2.1.2 Ground Based Hyperspectral Imaging System

For the outdoor ground hyperspectral data collection, we used a SOC710 Series® portable hyperspectral imager. The SOC710 imaging system was equipped with a 70 mm Xenoplan lens with a maximum aperture of $f$/8. The hyperspectral camera utilises a $512 \times 512$ pixel CCD sensor sensitive to radiation in the Visible to Near Infrared (VNIR), between 450 to 950 nm. Individual images are produced for each one of the 240 different bands covering the spectrum, with a dynamic 12-bit range [5]. The camera was connected to a PC laptop and operated by means of *Lumenera* drivers v.6.3.0 provided by the manufacturer.

### 2.1.3 Hyperspectral Data Collection

In Fig. 3, a true color rendering of the calibration and scene setup is shown as imaged by the SOC 710 from an elevated position. The setup consisted of a group of spectral targets (e.g., a soccer ball, pieces of clothing, artificial plant, a piece of paper carton), a white tarp with printed text in different colors and bar/checkerboard patterns which can be used to characterize spatial response function of an imager. Below the white tarp a tarp with three different reflection levels was positioned which allows to perform a simple gain calibration of the measured Digital Numbers (DN) of the tarp ($DN_{Tarp}$) as shown in Eq. (1). The more commonly used "empirical line method" was not used since the offset was assumed to be negligible and to avoid introducing more noise. A portable field spectrometer from ASD Inc. [6] was available to measure absolute reflectances of the three grey level target using a Spectralon© calibration panel from 400 to 2500 nm in 1 nm spectral resolution. In order to measure the best signal, the brightest tarp segment was used and a rectangular region of interest was used to calculate the mean DN for each band.

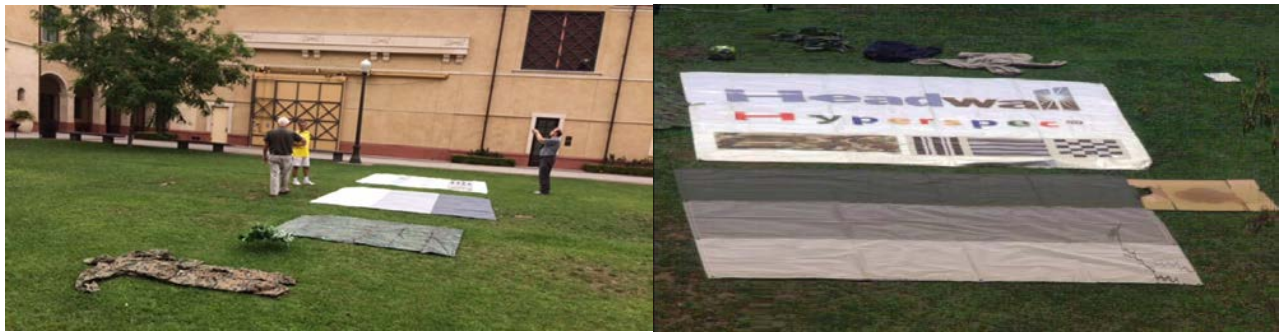

**Figure 3: Calibration and scene setup for the ground hyperspectral data collection at USC main campus.**

The conversion from DN to reflectance ($r_{SOC}$) using the SOC hyperspectral sensor is attained from

$$r_{SOC} = DN_{SOC} \cdot \frac{r_{ASD}}{DN_{Tarp}}, \qquad (1)$$

where $DN_{SOC}$ denotes the average DN of a pixel on the desired material from the SOC data cube, $r_{ASD}$ is the wavelength resampled averaged ASD absolute reflectance spectrum of the tarp measured by the ASD spectrometer, and $DN_{Tarp}$ is the average over a Region of Interest (ROI) of the tarp measured by the SOC hyperspectral sensor.

Fig. 4 plots the measured absolute reflectance spectra for the ASD spectrometer (Fig. 4 (left)), the SOC DN's for a single pixel (Fig. 4 (center)), and the computed SOC reflectances (Fig. 4 (right)) as a function of the SOC wavelengths. Notice the increasing noise for wavelengths above 800 nm which is due to fall-off in the sensitivity of the silicon based detector. The original plan was to fly the Nano-Spec hyperspectral imager on the Aibot UAV and image the scene at the same time with the ground based imager. However due to technical difficulties only ground-based data was collected for this site. A very similar setup was used later in a different location on top of a parking garage to calibrate the Nano-Spec imagery obtained from the Aibot small drone.

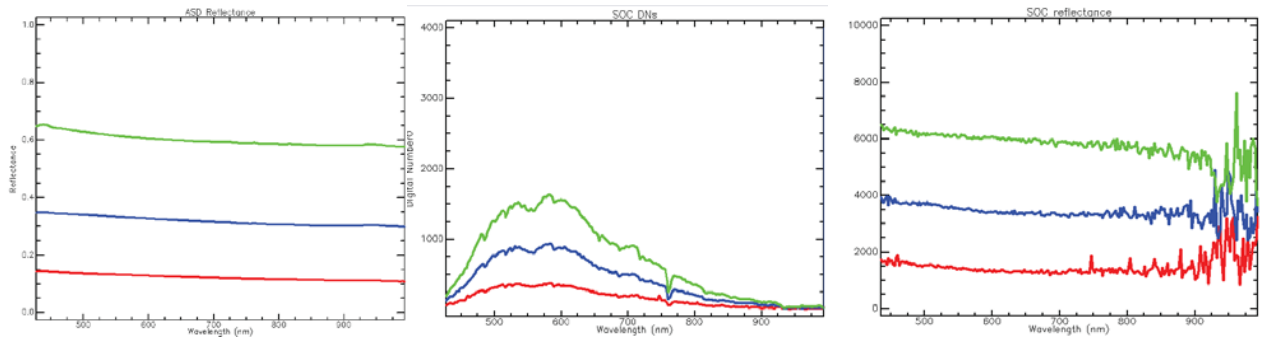

**Figure 4: Ground collection spectra: ASD tarp spectra (left), tarp DNs (center), SOC tarp reflectance spectra (right).**

Using the same calibration procedure for the Nano-Spec raw DN, data spectra shown in Fig. 5 were obtained using average DN's over a region of interest (Fig. 5 (center)). The Nano-Spec radiance data shows the oxygen absorption at 760nm and water vapor at 940 nm very clearly. Notice that the retrieved reflectances (Fig. 5 (right)) does not show a significant amount of noise when averaged over a region. However in Fig. 6, one can see large variation and increasing noise over relatively uniform road and roof areas, Fig. 6 (center plots) and Fig. 6 (right plots), respectively; where the black curves denote the mean, purple indicates +/- one standard deviation around the mean, and the red curves show the absolute maximum or minimum values. We have no explanation for the sinusoidal feature near 760 nm in the vegetation spectrum, which to the best of our knowledge in remote sensing theory it should not have be there.

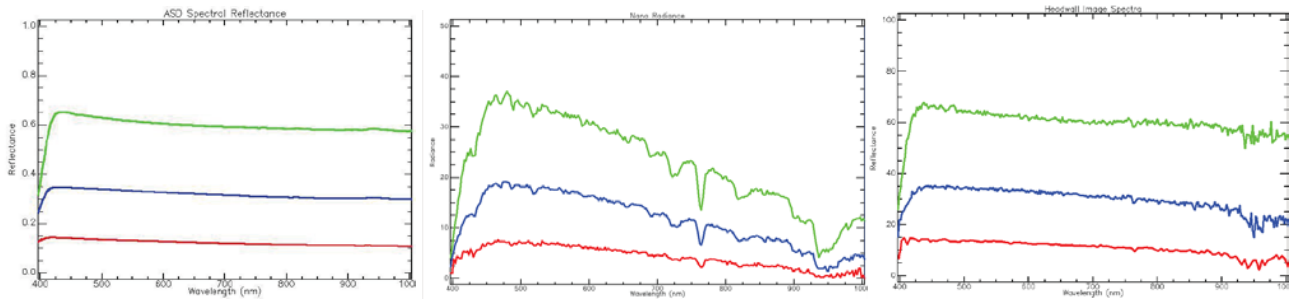

**Figure 5: Aerial collection spectra: ASD tarp spectra (left), Nano tarp radiance spectra (center), Nano tarp reflectance spectra (right).**

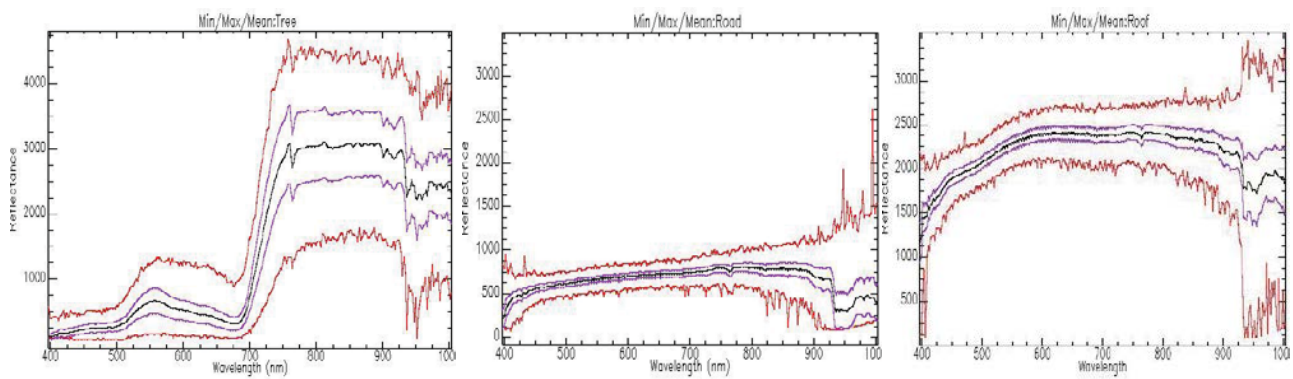

**Figure 6:** B**asic spectral statistics of key material classes: Tree (left), Road (center), and Roof (right).**

Fig. 7 shows a true color (left image) and false color (right image) composite of a geo-registered Nano-Spec data cube over the parking garage, which includes the calibration tarp, some trees and a road. The false color image highlights the value added for using spectral properties of specific material types to produce image segmentation maps of meaningful material types in the scene, while bypassing the requirement for machine learning.

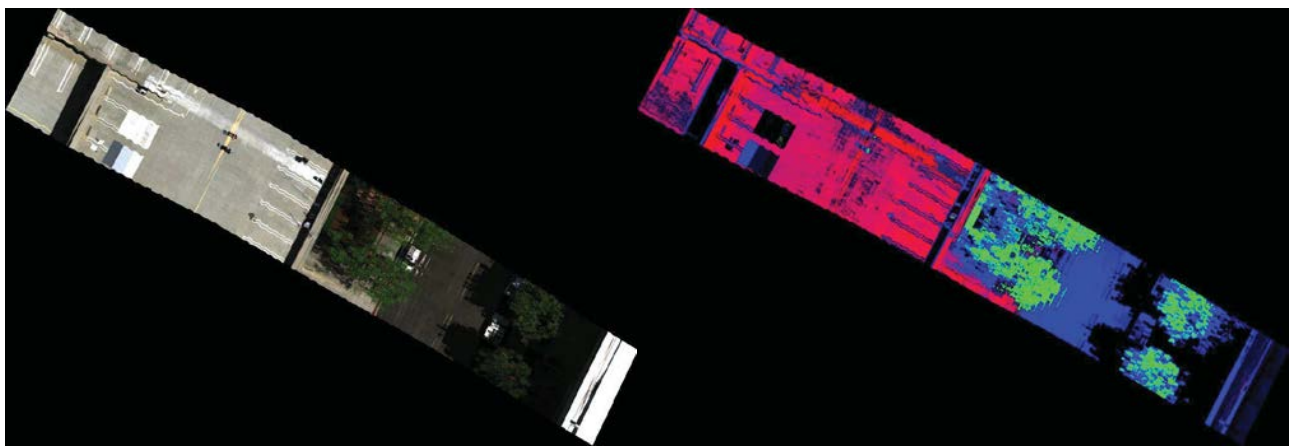

**Figure 7: (Left) Nano hyperspectral data cube example (true color (left)). (Right) Material map based on spectral property: Paved road (Red), Vegetation (Green), and Shade (Blue).**

## 3.0 DATA COLLECTION FOR PHOTOGRAMMETRY

3D models of terrain can be developed from many sources, including airborne or ground-based LIDAR, manual creation by artists, or photogrammetry. Photogrammetry, the 3D reconstruction of an area from images, provides advantages relative to LIDAR, which is often more expensive, time-consuming, and physically more demanding. In photogrammetry, a structure-from-motion (SFM) algorithm is used to reconstruct the geometry of a static scene, given a moving camera perspective. Photogrammetry does not require emitting energy (unlike LIDAR), which offers obvious advantages in the context of defense missions, as well as reduced power and weight requirements. Photogrammetry requires at least three images of a given feature from different positions and orientations to reconstruct that point in 3D space (Figure 8). Industry guidelines [7] suggest that images used for photogrammetry should have 60% or greater overlap to maximize the quality of the reconstruction. Since the 3D structure of the terrain and objects is reconstructed entirely from the images, it is necessary to have coverage of any part of the environment that needs to be included – algorithms cannot generate surface detail for areas that are not imaged from at least two (preferably many more) perspectives. The resolution of the final reconstruction is a function of the sensor system resolution, field of view, and platform altitude above the surface to be reconstructed, and results in tradeoffs between speed of data acquisition and detail level of the reconstruction.

Contemporary drone platforms, such as the DJI Phantom 3 and DJI Phantom 4, include integration with the imaging system and gimbal, allowing the development of flight plans that specify geographic waypoints, gimbal pitch, and imaging requirements (ISO, white balance, shutter timing). DJI also offers a Mobile Software Development Kit (SDK) that permits developers to interact with the drone platform via Android or iOS mobile devices. Similar SDKs exist for other control systems such as MAVLink, an open drone control protocol used by many open-source drone autopilot systems. Commercial photogrammetry applications exist for these systems, which produces a 3D dense point cloud of the imaged scene.

A dense 3D point clouds is a data structure containing information about a collection of points in three-dimensional space. The point cloud does not contain data about any surface that may be defined by the points, but may contain additional per-point information such as color. To generate a point cloud, a large corpus of photos must be aligned, and a dense point cloud generated from this data (Fig. 9). Existing commercial and research tools (Agisoft Photoscan, Pix4D Mapper) provide the necessary algorithms to process a set of overlapping two-dimensional visible-spectrum photographs into a high-density point cloud. These tools implement similar pipelines from a set of images to a dense point cloud. Each photograph is analyzed individually to identify recognizable image features. An algorithm such as Scale Invariant Feature Transforms (SIFT) or Speeded Up Robust Features (SURF) [8] is generally used in this step, since these algorithms produce feature descriptors that remain consistent across images when the camera moves relative to the target scene. These image features points are then correlated across multiple images. Similarity metrics are used to determine when a feature point appears in multiple images. External data including geotags embedded in image EXIF metadata may be used to constrain the search space for correlation. These correlated points are termed "tie points," and successful point cloud reconstruction requires on the order of tens of thousands of tie points. Once tie points are established, the three-dimensional location and orientation of the camera for each photograph is determined using a bundle adjustment algorithm. This algorithm attempts to optimize the placement of each camera position such that error across all tie points is minimized. The aligned images and known tie points are then used to reconstruct a dense point cloud. This process fills in the majority of spatial data by projecting back through the camera model to determine the depth of each pixel in each source image. The output of this pipeline is a dense point cloud and a set of aligned photographs.

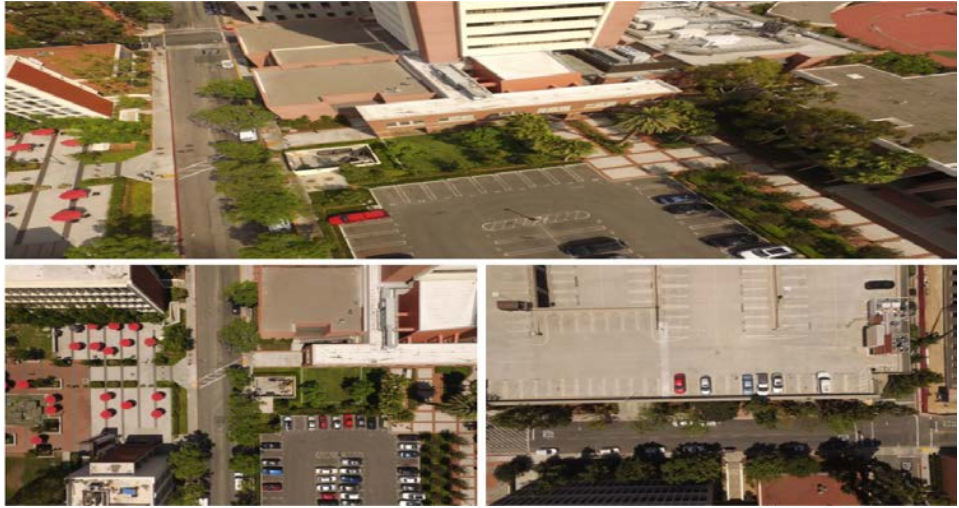

**Figure 8: Multiple aerial photographs of a pedestrian overpass. These images are representative; over 100 photos including this structure contributed to the reconstruction.**

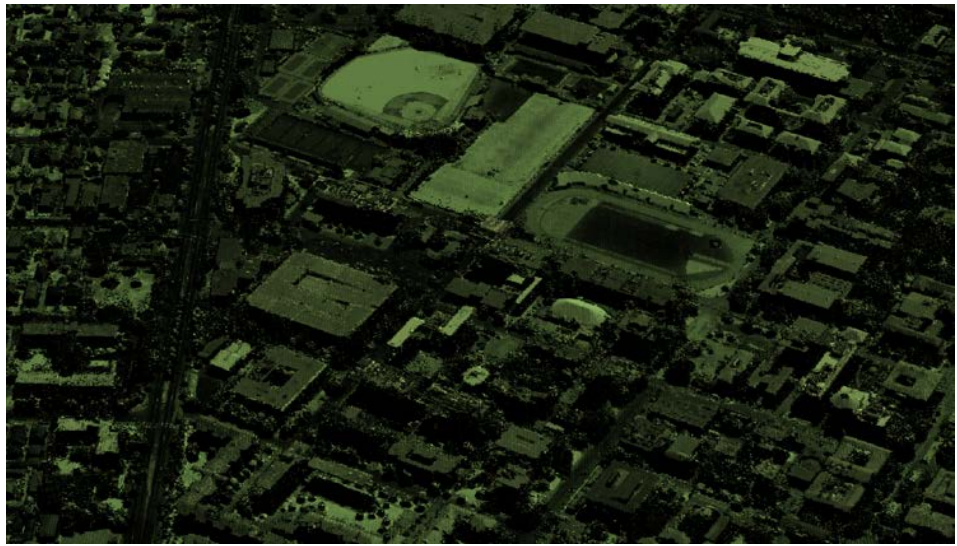

**Figure 9: Photogrammetric reconstruction of a portion of USC main campus, illustrating the dense point clouds.**

## 4.0   FUSION OF 3D RECONSTRUCTED AND HYPERSPECTRAL DATA

Inspired by image-based localization approaches similar to the one introduced in [9], we proposed an approach to register the 2D spatial area of hyperspectral data cubes onto the available 3D reconstructed models of the same scene, producing in the process the association of height information with each spectra. The approach follows these major steps:

(a) The fusion process starts by applying the K-means clustering algorithm to cluster all points in the 3D model's point clouds into k clusters. We arbitrarily used in this experiment 100,000 clusters, given the size of the point clouds.

(b) To improve computational performance, while facilitating solution convergence, we avoid comparing all of the points in the point clouds by assigning the centroids obtained from the k-means to be the *visual words*.

(c) Since Scale Invariant Feature Transform (SIFT) [10] was used as part of the 3D reconstruction step, SIFT is therefore justifiably applied to the RGB representation of hyperspectral data cubes, and also represented as visual words for comparison with visual words from (b).

(d) The visual word comparison strategy can then be reduced by finding through a linear search the two nearest visual words in (b) to the visual word (hyperspectral-based SIFT descriptor) in (c), using the K-Nearest Neighbor (KNN) algorithm and the smallest Euclidian distance with a user set minimum threshold; this is accomplished by first building a k-dimensional tree (k-d tree) for K set to 100,000 visual words. (K-d tree is a space-partitioning data structure for organizing points in a k-dimensional space. K-d trees are a useful data structure for several applications, such as searches involving a multidimensional search key, e.g., nearest neighbor searches.)

(e) The linear search in (d) yields as set of what will be referred to herein forth as *correspondences* between each hyperspectral-based SIFT descriptor and two 3D model visual words. The linear search continues until a user specified number of correspondence is satisfied (in this experiment this number was set to five), or the search is exhausted from the hyperspectral data perspective.

(f) The correspondence set attained in (e) does not necessarily guarantee a geometric alignment of the hyperspectral images onto the 3D model that would match the quality achieved by a subtle manual help from a human. So, in order to improve further the autonomous alignment process, we propose applying the Optimal Randomized RANSAC algorithm to the set of correspondence vectors obtained in (e), using the criterion that if more than $n$ correspondences ($n = 5$ worked well in this experiment) are inline, the geometric alignment is declared as acceptable and hence validated, where individual spectrum can then be assigned to height information from the 3D model.

The optimal value of the decision threshold is guaranteed by two theorems proved in [11]: Theorem 1 - The probability of rejecting a *good* model, and Theorem 2 (Wald's lemma) - The average number of observations (checked data points) carried out while testing a *bad* model can be derived in closed form. Fusion is complete.

The method described in this section was coded and automatically applied to the datasets described in Section 2, successfully and efficiently matching key landmarks between the 2D spatial representations of hyperspectral data cubes and the reconstructed 3D models of the target area in the USC main campus, as shown in Fig. 10.

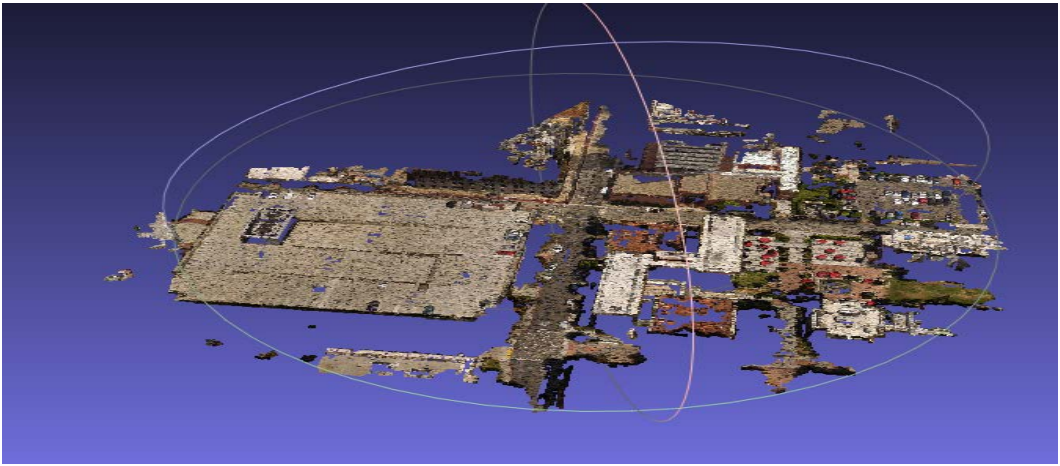

**Figure 10: Fusion of hyperspectral data with 3D model; geo-registration information was not required.**

The result depicted in Fig. 10 allows us to make the desired association between spectra and height information of the digitized target scene, although in that figure only the three band (red, green, blue) colors are shown being associated with 3D points in the scene. The approach is able to accomplish the registration task, without relying on the data's geo-rectified information that may or may not be available with the dataset. This is an enabling capability.

Fig. 11 depicts the mosaic registration of multiple hyperspectral data cubes, representing the target area at USC main campus, but using the geo-rectified data information of both datasets (3D digital elevation model and hyperspectral data), which is easier to accomplish.

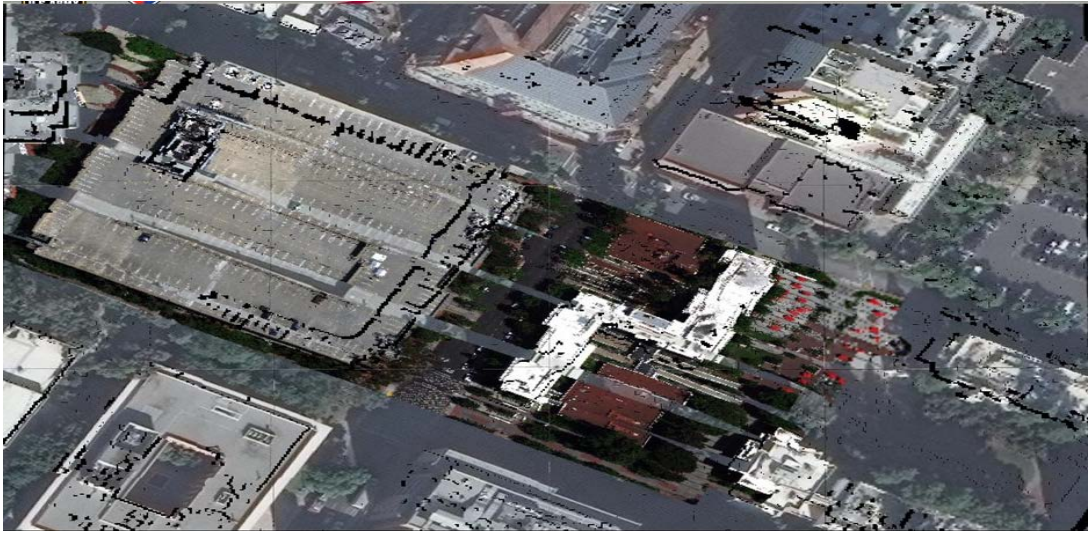

**Figure 11: Fusion of hyperspectral data with 3D model; geo-registration information was required.**

The drawback on results shown in Fig. 11 is the approach's dependence on Global Positioning Service (GPS), GPS receivers, and Inertial Measurement Unit (IMU), which may be denied in certain geo-locations in the world or simply unreliable.

It is worth noting that the fusion results presented in this paper can improve the user's situational awareness by post-processing the association between spectral and elevation data, albeit details on producing the target area's material segmentation maps are beyond the scope of this paper.

## 5.0 CONCLUSION

We discussed a field experiment conducted at the USC main campus, which employed COTS aerial drone solutions for data acquisition with compact broadband and hyperspectral sensors, and the details on our approach for data calibration and analysis, production of photogrammetry based 3D models, and fusion of hyperspectral data with the recovered set of 3D point clouds of a target area. Our research continues aiming for improving the machine's accuracy on fully automatic scene segmentation into key material classes for applications of interest to the NATO research community. Results from this experiment also supported an initiative formalized in 2016 between ARL and USC ICT through a summer intern mentorship program at ARL West.

The motivation for this field experiment and subsequent research were drawn from the fact that smaller cameras, better optics, cheaper cost, more efficient batteries, increased stabilization, intelligent image capturing, and the ability to include all of these capabilities on fully-autonomous craft (air or ground) affords the research community a wonderful opportunity to examine new ways of producing usable datasets and associated processing and algorithms for improved situational awareness. Combine these facts with advances in small drone design and availability, and other sensors that move beyond broadband cameras in the visible spectrum (e.g., hyperspectral), and there is little debate that the quality and quantity of data to produce a usable 3D geo-specific dataset of particular areas of interest in the world will increase in the coming years.

In this paper, we have presented only one possible approach to exploiting these capabilities: bringing together both the 3D scene reconstruction capability from images and fusion of material surface spectra with recovered elevation data from images. Our 3D point clouds and hyperspectral data fusion approach leveraged multiple existing methods in the 3D reconstruction step of the overarching solution, to include SIFT [10] and incremental structure from motion via Snavely Bundler technique to recover the 3D structure of the scene, using as input a large set of overlapping images from the scene. Motivated by the image-based localization method by Sattler et al. in [9] and the Optimal Randomized RANSAC algorithm [11], we proposed a method to register 2D spatial information from hyperspectral data with a set of 3D recovered points of the same area, and yield in the process the desired association between elevation and material spectra, to include values in the red, green, and blue (RGB) wavelengths, for further material classification processing. The approach is able to accomplish the registration task, without relying on the dataset's geo-rectified information that may not be available for all datasets. This is an enabling capability.

We hypothesize that the fusion of material spectral characteristics and recovered surface elevation data from images can significantly improve the reliability of a machine automatically generating material segmentation maps. We further hypothesize that such an improvement will likely augment the situational awareness of the user, and enable adaptive machine learning in applications requiring material target detection and identification. Results on these applications are beyond the scope of this paper, albeit the interest is noted for a future paper.

## 6.0 REFERENCES

[1] ARL West website's article (Date last accessed: 5 Mar 2017): [https://www.arl.army.mil/www/?article=2756]

**Small Drone Field Experiment: Data Collection & Processing**

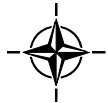